\documentclass[runningheads]{llncs}
\usepackage{graphicx}
\usepackage{amsmath,amssymb} % define this before the line numbering.
\usepackage{color}
\usepackage{xspace}
\usepackage{subfig}
\newcommand{\etal}{\emph{et al. }}
\usepackage{xcolor}
\usepackage{float}
\usepackage{multirow}
\usepackage{flexisym}
\usepackage[linesnumbered,ruled,vlined]{algorithm2e}
\usepackage{algpseudocode}
\renewcommand{\algorithmiccomment}[1]{}

\begin{document}
% \renewcommand\thelinenumber{\color[rgb]{0.2,0.5,0.8}\normalfont\sffamily\scriptsize\arabic{linenumber}\color[rgb]{0,0,0}}
% \renewcommand\makeLineNumber {\hss\thelinenumber\ \hspace{6mm} \rlap{\hskip\textwidth\ \hspace{6.5mm}\thelinenumber}}
% \linenumbers
\pagestyle{headings}
\mainmatter
\def\ECCV18SubNumber{9}  % Insert your submission number here

\title{MASON: A Model AgnoStic ObjectNess Framework}

\titlerunning{MASON: A Model AgnoStic ObjectNess Framework}
\authorrunning{K J Joseph
% , Rajiv Chunilal Patel 
and Vineeth N Balasubramanian}
% Sir, I was referring to https://cs.stanford.edu/people/jcjohns/papers/eccv16/JohnsonECCV16.pdf for an example of authorrunning

\author{K J Joseph
% \and Rajiv Chunilal Patel\inst{2} 
\and Vineeth N Balasubramanian}

% Replace with shorter version of the author list. If there are more authors than fits a line, please use A. Author et al.
%

\institute{Indian Institute of Technology, Hyderabad, India \\ \email{\{cs17m18p100001,vineethnb\}@iith.ac.in} 
% \andANURAG, Defense Research and Development Organization, India\\
% \email{rc\_patel@anurag.drdo.in}
}

\maketitle

\newcommand{\method}{MASON\xspace}

\begin{abstract}
This paper proposes a simple, yet very effective method to localize dominant foreground objects in an image, to  pixel-level precision. The proposed method `MASON' (Model-AgnoStic ObjectNess) uses a deep convolutional network to generate category-independent and model-agnostic heat maps for any image. The network is not explicitly trained for the task, and hence, can be used off-the-shelf in tandem with any other network or task. We show that this framework scales to a wide variety of images, and illustrate the effectiveness of MASON in three varied application contexts.
\end{abstract}

\section{Introduction}
Identifying the pixels in an image that contribute towards the most distinct object(s) in it, is an important computer vision task. This forms a core component of many downstream tasks such as object detection, instance segmentation and object tracking to name a few. Traditional methods for foreground segmentation use texture information \cite{karoui2010variational}, edge information \cite{mortensen1995intelligent} or graph-cut based methods \cite{boykov2001interactive,grabcut} to infer the location of the foreground objects. Some of these methods enable users to interactively provide suggestions to modify the segmentation results, thus allowing for better results at the cost of manual intervention. 

Convolutional Neural Networks (ConvNets) have shown superior performance in image classification \cite{krizhevsky2012imagenet,resnet,vgg-16}, object detection \cite{fasterrcnn,rfcn}, semantic segmentation \cite{deeplab,long2015fully} and many other computer vision tasks. It is widely understood now that the feature maps in consecutive layers of ConvNets capture a hierarchical nature of the representations that the network learns. This typically starts with detectors for edges with different orientations or blobs of different colors in earlier layers, followed by progressively learning more abstract features. An extensive analysis of these hierarchical representations was presented by Zeiler and Fergus in \cite{zeiler2014visualizing}. 

%%%%%%%%% FIGURE STARTS
\begin{figure}
\centering
\subfloat[Input image]{\includegraphics[width=0.27\linewidth]{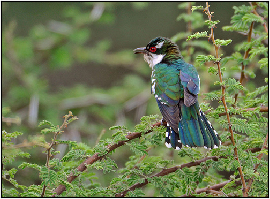}}\qquad
\subfloat[Generated heatmap]{\includegraphics[width=0.27\linewidth]{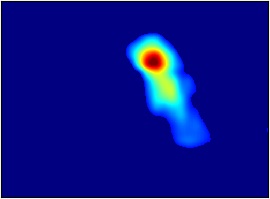}}\qquad
\subfloat[Segmentation result]{\includegraphics[width=0.27\linewidth]{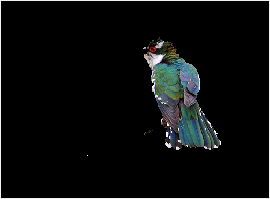}}\qquad
\caption{An illustration of the proposed method. Subfigure (a) is a test image from Pascal VOC 2012 \cite{pascal-voc-2012} dataset. Subfigure (b) is the objectness heat map that \method generates for the image. Subfigure (c) shows the segmentation result generated by following the methodology described in Section \ref{sec:IntegWithGrabCut} (best viewed in color).}
% * <vineethnb@iith.ac.in> 2018-07-21T00:43:47.720Z:
% 
% It is not clear from the caption what the model was trained on. Do you think we need to add that?  Or no?
% 
% ^.
\label{fig:our_method_illustrate}
\end{figure}
%%%%%%%%% FIGURE ENDS

In this work, we show that the feature maps (activations) from any of the convolutional layers in a deep network model, that is trained for image classification, contain sufficient information to pinpoint where exactly in a given image lies the dominant foreground object(s). This information can be used to construct a heat map, where the intensity at each location indicates the likelihood of the pixel representing a foreground object. We make use of this heat map in three different application settings. 
First, the heatmap acts as an effective alternative for the manual annotation required for interactive foreground segmentation algorithms to guide their segmentation. We find and show in this paper that this is a powerful combination that can provide impressive results even when the images contains multiple objects of different sizes. An illustration of the proposed method for foreground segmentation is shown in Figure \ref{fig:our_method_illustrate}. (We note that this result requires no additional training, given a pre-trained image classification model, and the method can be used as is for any input image.)

Next, we use the proposed framework to implement a generic method to extend existing object detectors to perform instance segmentation. Lastly, we make use of the proposed framework to improve the quality of poorly annotated object detection datasets by automatically filtering out false positive annotations and tightening bounding box annotations inscribing true ground truths. Our experiments validate that this pre-processing step helps the object detection algorithms to learn a more generic object detector on using the cleansed dataset. 

The remainder of this paper is organized as follows: We discuss literature related to our work in Section \ref{sec:RelatedWork}. The proposed Objectness Framework (\method) is introduced in Section  \ref{sec:ObjectnessFramework}, followed by detailed explanation of the aforementioned three application settings in Sections \ref{sec:IntegWithGrabCut} - \ref{sec:ImprovingDataset}. We showcase two additional capability of the method in Section \ref{sec:Discussion}. We conclude the paper with pointers to future work in Section \ref{sec:Conclusion}.

% ========RELATED WORK STARTS==========
\section{Related Work} 
\label{sec:RelatedWork}
We review the related literature from both recent work using ConvNets, as well as more traditional vision-based approaches that achieve a similar objective as our work.

\paragraph{ConvNet-based Object Localization:} 
Different methods with various levels of supervision have been explored in literature to localize objects in images. Convolutional Neural Networks have been used to predict the salient regions in images. The parts of an image that stand out from the background is often annotated in the ground truth in such a task. This ground truth is used to train a ConvNet, which learns generic features to predict the saliency of unseen images. \cite{Pan_2016_CVPR,saliency_2,saliency_3,saliency_4} work on this principle. These efforts show that they are able to generate better results when compared to  traditional vision-based saliency prediction methods.

A large number of successful semantic segmentation methods have been  based on Fully Convolutional Networks (FCNs) proposed by Long \etal in \cite{long2015fully}. SegNet \cite{DBLP:journals/corr/BadrinarayananK15} improved the performance of FCNs by using an encoder-decoder architecture followed by a pixel-wise softmax layer to predict the label at each pixel.  
U-Net \cite{unet} uses a fully convolutional architecture with a contraction and expansion phase, where the features from contraction phase is used in the expansion phase to recover the information lost while downsampling. 
DeepLab\cite{deeplab} introduces convolution with upsampled filters, atrous spatial pyramid pooling and a CRF based post processing to improve the segmentation benchmarks. 
The global context information available in the feature maps of different depths is utilized in PSPNet \cite{zhao2017pspnet}.
% * <vineethnb@iith.ac.in> 2018-07-21T00:37:50.043Z:
% 
% Looks like DeepLab and PSPNet are glaringly missing from the semantic segmentation survey, Joseph. Adding a line on each would be great (just to not allow the reviewer to point this out)
% 
% ^.
These methods need a large amount of data to train, which however can be offset to some extent by initializing the network using pretrained weights. A deeper issue in such methods is the need for large datasets which are densely annotated at the pixel-level, which is non-trivial and requires immense manual effort. In contrast, in this work, we propose to provide dense pixel-level segmentation of foreground objects with no explicit training of the network for the task. 

Segmenting an object from an image using its `objectness' has been explored over the years. Alexe \etal \cite{6133291} explored ways to quantify objectness and used it as the key cue to infer saliency in images in the Pascal VOC \cite{pascal-voc-2012} dataset.  However, such methods have largely been based on handcrafted features, which are fast losing relevance in this era of learned feature representations. The method proposed in \cite{pixelobjectness} generated an objectness map for foreground objects in images using a ConvNet that is trained on image-level and boundary-level annotations. Their method needs to be explicitly trained with pixel-level segmentation ground truth to achieve good performance, while we propose a framework that needs no explicit training for the segmentation task. 

Recently techniques like Class Activation Maps (CAM)\cite{zhou2016learning} and Grad-CAM \cite{Grad-CAM} has been proposed for producing visual explanation for the decisions from a large class of CNN models. They also use the activations from pretrained networks to generate heat-maps to localize the object in an image. While these methods localize objects of a specific class, our proposed method is class agnostic. We compare ourself with Grad-CAM in section \ref{sec:ComparisonWithCAM}. The results reveal that despite being much simpler than Grad-CAM, the method is able to achieve competitive results.

\paragraph{Traditional Vision-based Object Localization:}
Historically, vision-based methods for foreground segmentation have been studied for many years. Color, contrast and texture cues have been used often to segment dominant foreground objects. \cite{karoui2010variational} and \cite{mortensen1995intelligent} work on this principle. However, the more successful group of methods for foreground segmentation have been based on graph-cut methods. \cite{yi2012image} offers a holistic review on all graph-cut based algorithms for foreground segmentation. We leverage such methods to improve our object localization in this work.

Object proposal methods, that have been proposed in recent years as a pre-processing step for object detection, closely align with the motive of finding discriminative regions in an image. We compare our framework against methods of this kind, in particular: Edge Boxes \cite{EdgeBoxes}, Selective Search \cite{SelectiveSearch} and Randomized Prim's method \cite{RandomizedPrims} in Section \ref{sec:ComparisonWithOPM}.

% ========RELATED WORK ENDS==========

\section{\method: Proposed Framework} \label{sec:ObjectnessFramework}
We propose a simple but effective ConvNet-based method to find the location of those pixels that correspond to the most distinct object(s) in an image. We define this property of an image as its \textit{Objectness}.

%%%%%%%%% FIGURE STARTS
\begin{figure*}
\centering
\subfloat[Input image]{\includegraphics[width=0.1\linewidth]{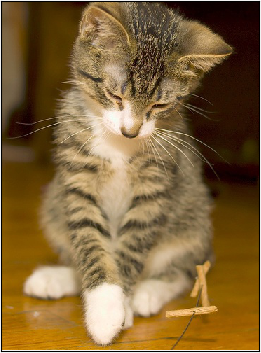}}\qquad
\subfloat[Top 63 features maps from conv5\_3 layer of VGG-16.]{\includegraphics[width=0.65\linewidth]{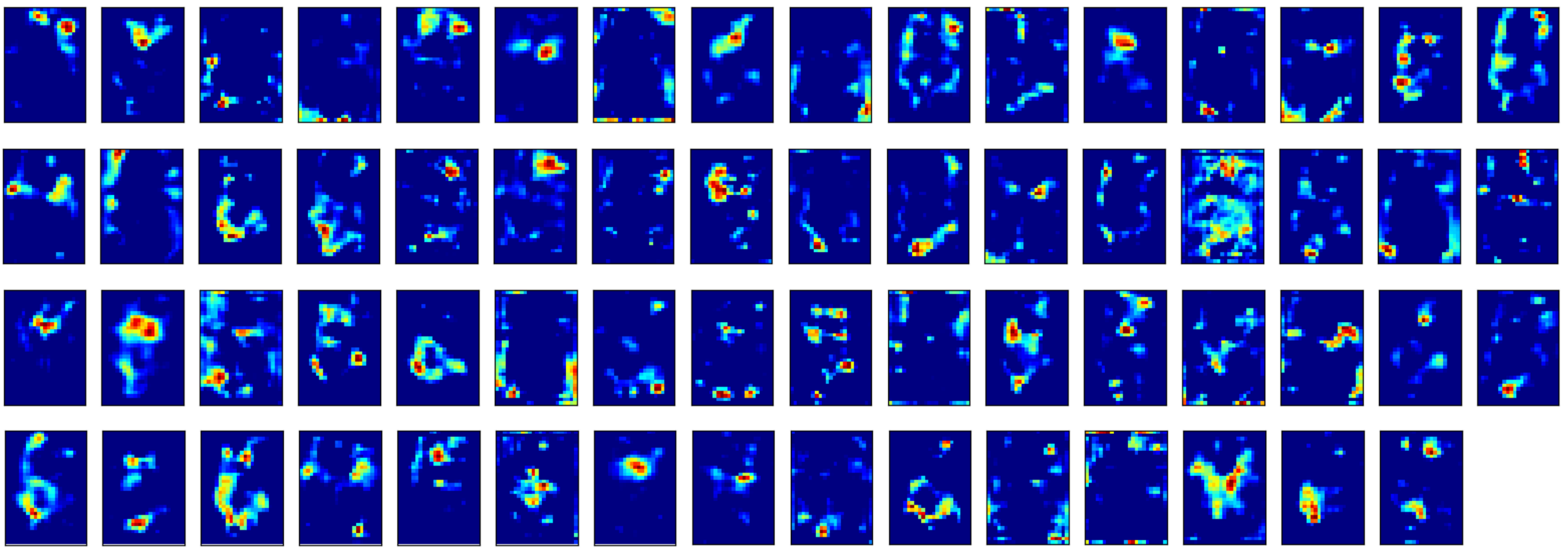}}\qquad
\subfloat[Objectness heatmap]{\includegraphics[width=0.1\linewidth]{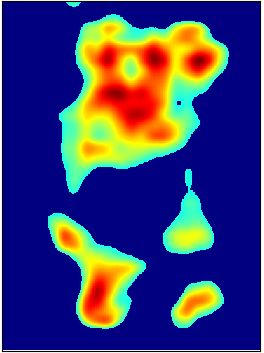}}\qquad
\caption{Subfigure (b) shows feature maps from conv5\_3 layer, when the Input image (Subfigure (a)) is passed through a VGG-16  \cite{vgg-16}, pretrained on ImageNet Dataset. Subfigure (c) is the generated Objectness heatmap. (best viewed in color).}
\label{fig:heatmaps_of_features}
\end{figure*}
%%%%%%%%% FIGURE ENDS

ConvNet architectures like AlexNet \cite{alexnet}, VGG-16 \cite{vgg-16} and ResNet \cite{resnet}, which are trained for the image classification task on large datasets like ImageNet \cite{imagenet_cvpr09} and PascalVOC \cite{pascal-voc-2012}, learn very good features in their kernels. The activations obtained from the different layers of these image classification networks, yield very good information for localizing the distinct object(s) in an image. 
Figure \ref{fig:heatmaps_of_features}(b) shows the top 63 feature maps (ranked by the  intensity of pixels) from a VGG-16 network trained on ImageNet, when the cat image in Figure \ref{fig:heatmaps_of_features}(a) is passed through it. It is evident that the activations provide strong cue about the location of the cat in the image, albeit distributed across the feature maps.
We exploit this information to localize objects from any given image. We hypothesize that a linear weighted combination of the activations from such networks (pre-trained on standard object recognition task), helps to generate a very useful heat map of the object in an image. In particular, in this work, we find that a simple sum of the activations provides useful results. We call this, the \textit{objectness heatmap}, $O$, of the object in the scene, i.e:
\vspace{-10pt}
\[O(I) = \sum_{i = 1}^{\left | f^{^{l}} \right |} f_{i}^{l}
\tag{1} \label{eq:1}
\]
% \vspace{-10pt}
where $I$ refers to the input image, $f^{^{l}}$ refers to the feature maps at layer $l$ and $f_{i}^{l}$ refers to $i^{th}$ feature map in layer $l$. The objectness heatmap, $O(I)$, that \method generates, is of the same dimensions as the input image, i.e. $O:\mathbb{R}^{m \times n} \rightarrow 
[0,1]^{mn}$. Each entry in $O(I)$ has a value which suggests the degree to which, the particular pixel contains an object or not.
%For those areas of the image which has high likelihood of having an object, the heat map will have high values. 
We observe that this sum of activations shows high likelihood at locations in the image with objects, and thus provides a useful objectness heatmap. (One could derive more complex variants by learning weights in a linear combination of the feature maps, but our experiments found the sum to work very well in practice.) An example is shown in Figure \ref{fig:our_method_illustrate} (b).

\begin{figure}[h]
\centering
\subfloat[Input Image]{\includegraphics[width=0.20\linewidth]{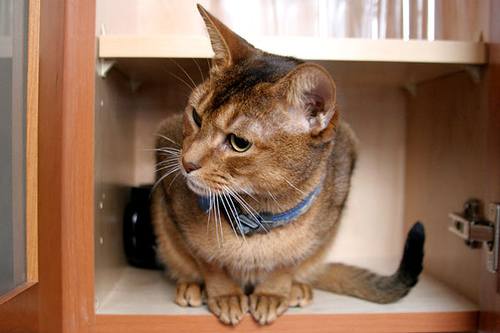}}\qquad
\subfloat[Heatmap from conv3\_3 layer of VGG-16.]{\includegraphics[width=0.20\linewidth]{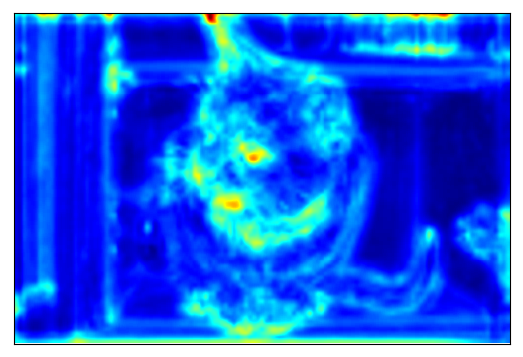}}\qquad
\subfloat[Heatmap from conv4\_1 layer of VGG-16.]{\includegraphics[width=0.20\linewidth]{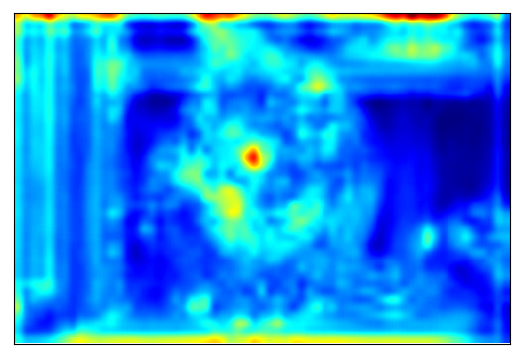}}\qquad
\subfloat[Heatmap from conv4\_2 layer of VGG-16.]{\includegraphics[width=0.20\linewidth]{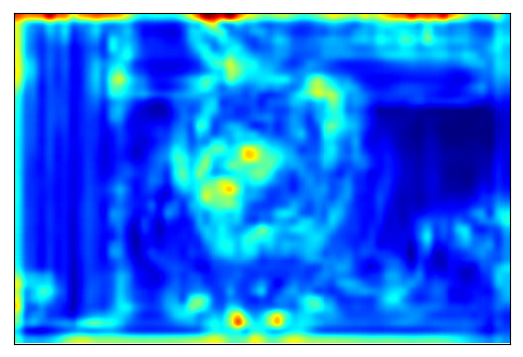}}\qquad
\subfloat[Heatmap from conv4\_3 layer of VGG-16.]{\includegraphics[width=0.20\linewidth]{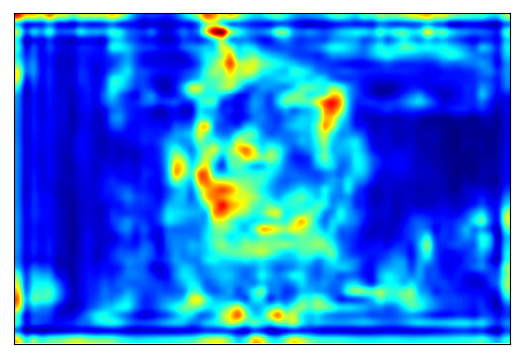}}\qquad
\subfloat[Heatmap from conv5\_1 layer of VGG-16.]{\includegraphics[width=0.20\linewidth]{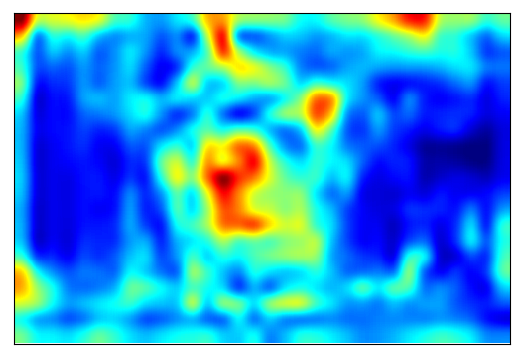}}\qquad
\subfloat[Heatmap from conv5\_2 layer of VGG-16.]{\includegraphics[width=0.20\linewidth]{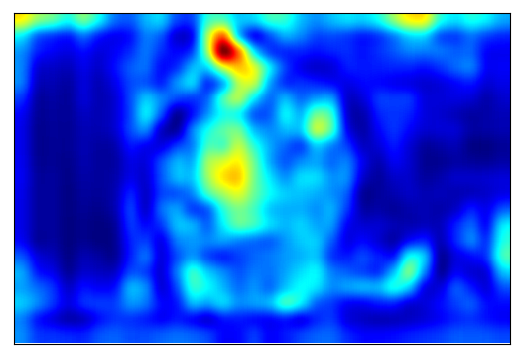}}\qquad
\subfloat[Heatmap from conv5\_3 layer of VGG-16]{\includegraphics[width=0.20\linewidth]{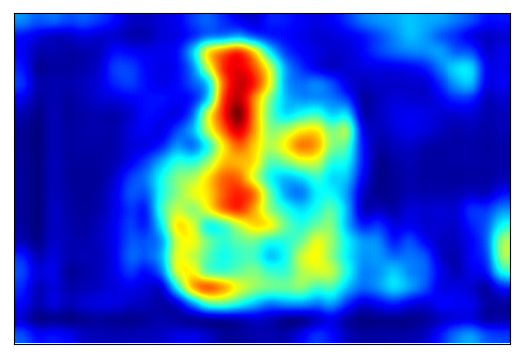}}\qquad
\caption{
The figure shows the different heatmaps generated by \method when feature maps of different layers of VGG-16 are used. Feature maps from deeper layers give a better cue for the location of the object rather than the earlier layers. 
% Heatmap generated by \method when the feature maps are taken from different layers of VGG-16. (best viewed in color)
}
\label{fig:maps2} 
\end{figure}

Despite its simplicity, this is an attractive option for finding the objectness of an image because this method can work out-of-the-box from the  pretrained image classification models available in public repositories. The only change that needs to be done is to remove the final fully connected layers. 
Through our experiments, we find that the features thus obtained generalize well to object categories beyond the ones they were originally trained for. Hence, we name our idea \textit{\method}: \textbf{M}odel \textbf{A}gno\textbf{S}tic \textbf{O}bject\textbf{N}ess framework. The ability to quantify the objectness in an image is generic across various ConvNet architectures that have been trained for image classification. In our evaluation with AlexNet \cite{alexnet}, CaffeNet \cite{caffe} and VGG-16 \cite{vgg-16}, the results consistently show the versatility of \method. In a given detector, the deeper the layer from which the activation maps are considered, the more is the granularity of the heat map. These visual results are shown in Figure \ref{fig:maps2}.

For all the results presented in this work, we use all  512 feature maps from conv5\_3 layer of VGG-16 network trained for image classification on the ImageNet dataset. Conv5\_3 layer is the last convolutional layer in the VGG-16 architecture. Owing to the high level features that are captured in these deeper layers \cite{zeiler2014visualizing}, these heat maps are much denser than heat maps from previous layers. The four pooling layers in the VGG-16 will successively reduce the size of the input image by half. Hence the resulting heat map will be 16x smaller than the input image. Bicubic interpolation is used to scale up the heat map to the original image size. The values of the heat map is scaled between 0 and 255.

We illustrate the usefulness of the \textit{objectness heatmap} in three application domains: (i) Task 1: Fine-grained object localization; (ii) Task 2: Extending object detectors to perform instance segmentation; and (iii) Task 3: Improving the quality of detection datasets. Algorithm \ref{algorithm2} provides an overview on how the objectness heatmap is used in each task. Section \ref{sec:IntegWithGrabCut} through \ref{sec:ImprovingDataset} explain each of these in detail.

\begin{algorithm}[H]\footnotesize
\SetAlgoLined
% \tcc{Comment}
\textbf{Input:} CNN model pre-trained for image classification $\Phi(.)$, Input Image $\textbf{I}$, Layer of interest $l$\\
\textbf{Output:} Binary object localization mask $O(\textbf{I})$ \\ 

\texttt{\\}
%\tcc{Generating the objectness heat map}
$f \leftarrow \phi(\textbf{I})$ \{Forward pass the image to generate the activations\}

$\tilde{O} \leftarrow \sum_{i = 1}^{\left | f^{^{l}} \right |} f_{i}^{l}$
\{Applying Equation \ref{eq:1} on features from layer $l$\}

$O(\textbf{I}) \leftarrow Bicubic\_Interpolation (\tilde{O})$ \{Interpolating to the original size\}\\
\texttt{\\}
\tcc{Task 1: Fine-grained object localization}
Stratify $O(\textbf{I})$ to obtain foreground and background\\
Use GrabCut \cite{grabcut} to generate fine-grained object localization\\

\texttt{\\}
\tcc{Task 2: Extending Object Detectors to Instance Segmentation}
For each predicted bounding box $\textbf{b} \in B$, stratify $O(\textbf{b})$ to obtain foreground and background\\
Use GrabCut \cite{grabcut} to generate instance segmentation in \textbf{b}\\

\texttt{\\}
\tcc{Task 3: Improving Quality of Detection Datasets}
For each predicted bounding box $\textbf{b} \in B$, compute $O(\textbf{b})$\\
Crop bounding box to box inscribing largest contour of $O(\textbf{b})$\\

\caption{\method Methodology}
\label{algorithm2}
\end{algorithm}

%%%%%%%%% FIGURE ENDS

\subsection{Fine-grained Object Localization using \method} \label{sec:IntegWithGrabCut}

The objectness heatmap $O(I)$ (Eqn \ref{eq:1}) obtained using \method can be combined with GrabCut \cite{grabcut}, a graph cut based foreground extraction method, to generate fine-grained object localization in an image. GrabCut lets the user specify one bounding box for the object under consideration as well as pixel-wise `strokes' which explicitly specifies those regions in the image which is surely a foreground or a background. 
A Gaussian Mixture Model (GMM) is used to model the foreground and the background, whose initialization is dependent on the user specified cues. A graph is built from this pixel distribution where pixels are considered as vertices. Each foreground pixel is connected to a pseudo-vertex called Source and the background pixels to another pseudo-vertex called Sink. The weight of the edges that connects the source/sink nodes to their respective pixels is defined by the probability of that pixel being foreground or background. The weight of edges between pixels is proportional to inter pixel similarity. Then a min-cut algorithm is used to segment the graph. The process is iterated until the GMM converges. 

\begin{figure}\centering {\includegraphics[width=0.95\textwidth]
{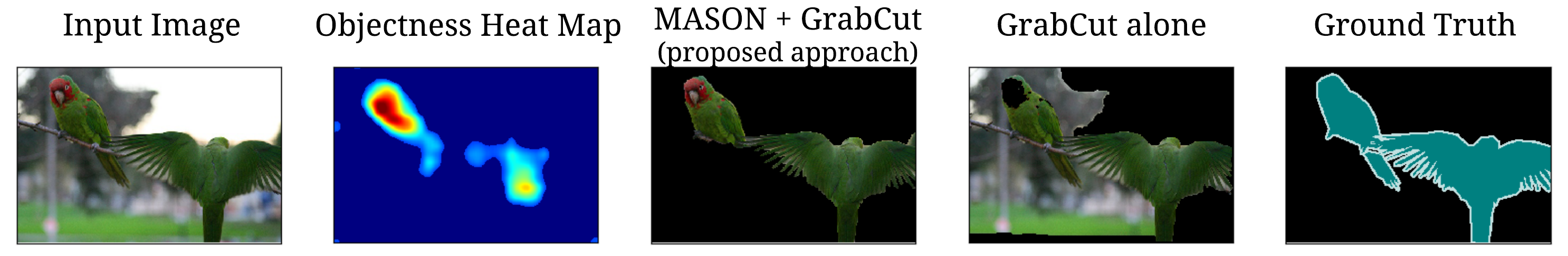}}
{\includegraphics[width=0.95\textwidth]{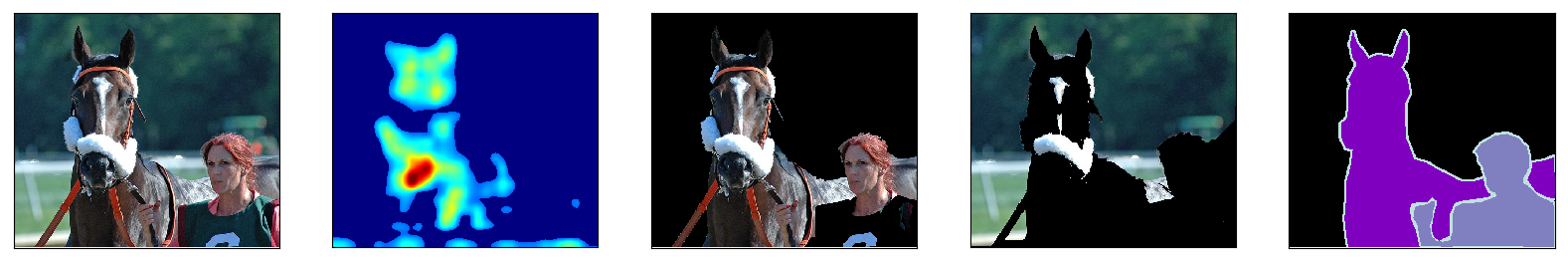}}
{\includegraphics[width=0.95\textwidth]{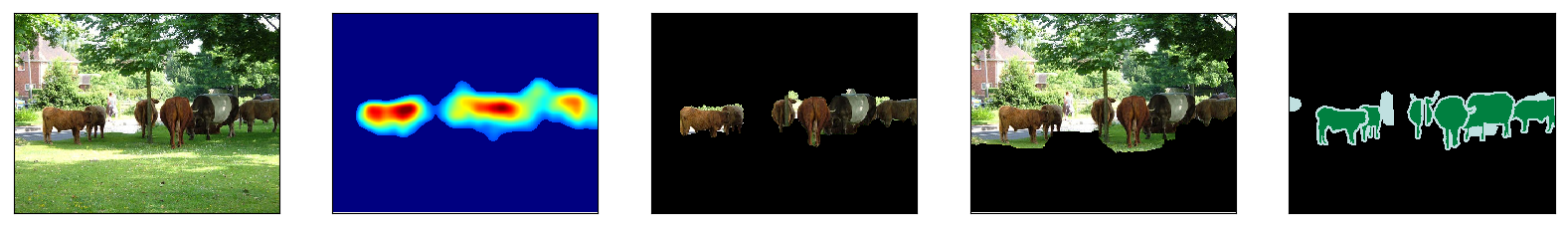}}
{\includegraphics[width=0.95\textwidth]{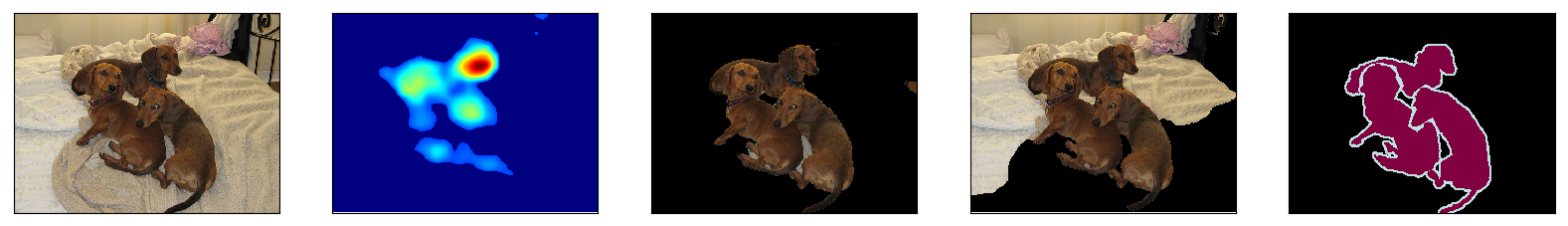}}
{\includegraphics[width=0.95\textwidth]{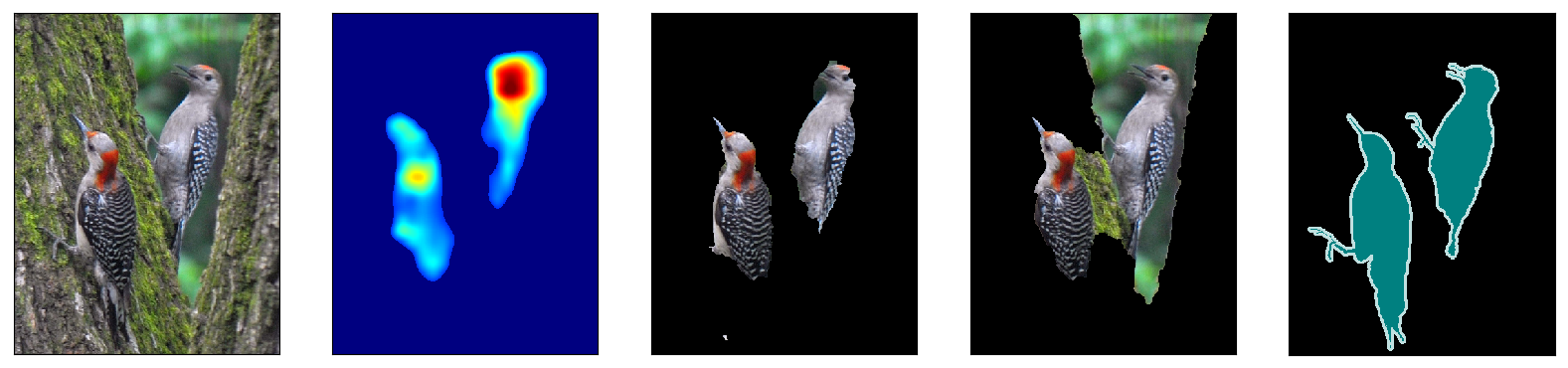}}
{\includegraphics[width=0.95\textwidth]{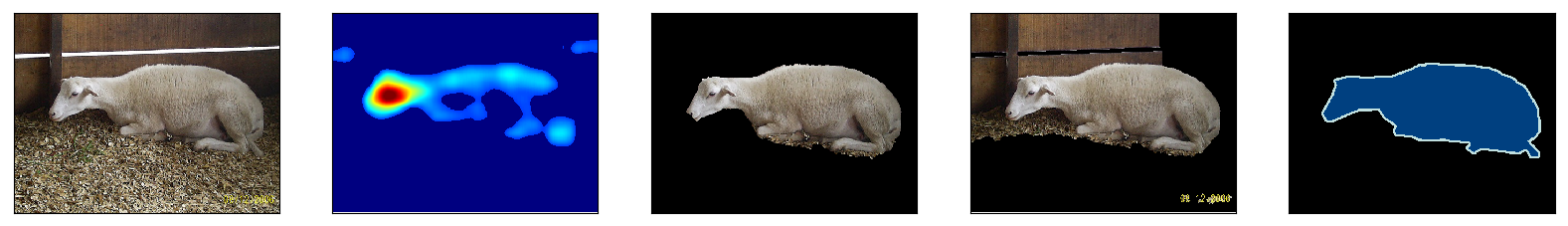}}
{\includegraphics[width=0.95\textwidth]{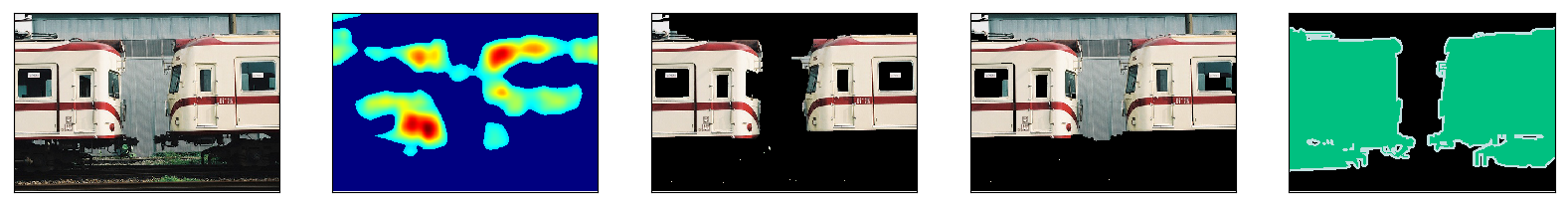}}
{\includegraphics[width=0.95\textwidth]{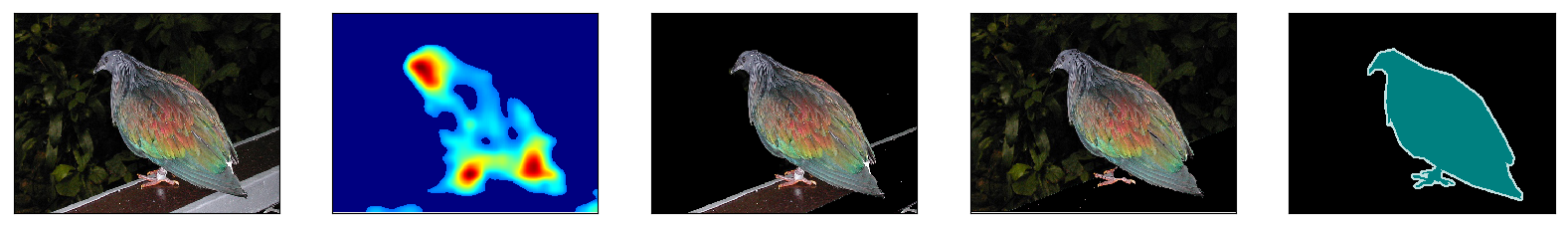}}
{\includegraphics[width=0.95\textwidth]{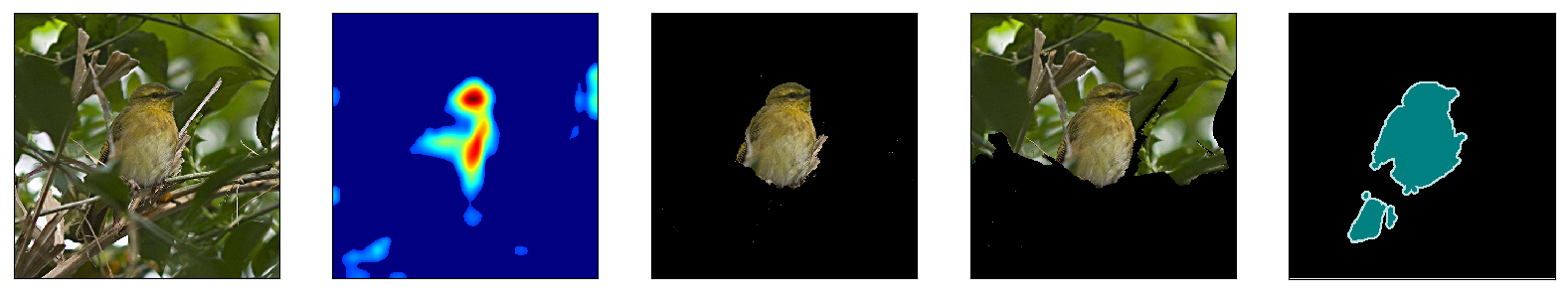}}
    \caption[]{
    The effectiveness of \method in fine-grained object localization is captured in the figure (Column 3). The proposed method (Column 3) is more effective than just using Grabcut \cite{grabcut} (Column 4) in foreground segmentation.
% More visual examples are in Supplementary Section. 
%     (best viewed in color).
}
    \label{fig:segmentation_results}
\end{figure}

% \textbf{Combining \method with GrabCut} is very ideal as both the methods can build on the strengths of each other.
In this work, we stratify the pixel intensities from the objectness heatmap to automatically generate the foreground and background `strokes' for GrabCut, and show that the results thereby generated are very informative.
In particular, we specify the `strokes' as a two-dimensional mask on a given image with each entry in the mask as one of four values; 0 (sure background), 1 (sure foreground), 2 (probable background), and 3 (probable foreground). All regions of the objectness heatmap generated by \method with pixel intensity greater than the mean intensity are labeled as 1 (sure foreground) in the mask. Those regions whose intensity is between zero and the mean intensity is labeled 3 (probable foreground) and those with zero intensity value are labeled 2 (probable background). The additional information that \method provides, enables GrabCut to perform foreground extraction well, even when there are multiple objects in the scene (Figure \ref{fig:segmentation_results}), which is beyond the scope of the vanilla GrabCut algorithm. 

% Horizontal Image STARTS
% Horizontal Image ENDS
This methodology helps generate good foreground segmentation for any image without any explicit training. 
% This process is unsupervised; as neither \method nor GrabCut has had any explicit training to generate  segmentation. 
The ConvNet that powers \method has been trained for image classification on the ImageNet \cite{imagenet_cvpr09} dataset, and doesn't need finetuning to generate the objectness map. Besides, GrabCut does not need training too. We evaluate the segmentation result on Pascal VOC 2012 \cite{pascal-voc-2012} test set. Each image from the test set is passed to \method and the segmentation that is generated is compared with the ground truth. Mean IoU is used as the evaluation metric.  Our proposed foreground segmentation method is able to achieve 0.623532 mean IoU with the ground truth. Figure \ref{fig:segmentation_results} shows example qualitative results of \method's performance. 
% More visual examples are provided in the Supplementary Section.

\subsection{From Object Detection to Instance Segmentation using \method}\label{sec:RepurposingDetectors}
ConvNet based Object Detectors like Faster-RCNN \cite{fasterrcnn}, R-FCN \cite{rfcn}, YOLO \cite{yolo} and SSD \cite{ssd} have shown competent performance in detecting objects from real-world images in MS COCO \cite{coco} and Pascal VOC \cite{pascal-voc-2012} challenges. We show how \method can be used to extend the capability of an existing object detector model to perform instance segmentation of the detected objects, without additional training. 

Any object detector takes an image as input and produces a set of bounding boxes with the associated class label as its output. Each of those areas that are inscribed by the corresponding bounding boxes can be passed through the \method+ GrabCut method as explained in Section \ref{sec:IntegWithGrabCut}, to obtain the foreground segmentation for that small area. This is combined with the class label information that the object detector predicts to generate instance segmentation for the objects that are detected by the object detector. For those detections that overlap each other, the segmentation with lesser area is overlaid on top of the others.

While the methodology that is proposed above can work for any object detectors, region based object detectors like Faster-RCNN \cite{fasterrcnn}, R-FCN \cite{rfcn} etc. can be modified without any retraining to integrate \method into its architecture. This will enable such object detectors to produce instance segmentations in a nearly cost-free manner. The family of region based object detectors has a classification head and a regression head, which shares most of the computation with a backbone CNN. Image classification networks architectures like VGG \cite{vgg-16} or ResNet \cite{resnet} is usually used as the backbone CNN. Its weights are initialized with that of ImageNet \cite{imagenet_cvpr09} training. During the joint training of the object detector, the weights are altered. Still, we find that the backbone CNN retains the property to generate Objectness heatmap with significant detail. This heat map can be used to generate foreground segmentation following the further steps explained in Section \ref{sec:IntegWithGrabCut}, which in-turn can be combined to generate instance segmentation for the image.

\begin{figure}[h]
\centering
% \subfloat[]{\includegraphics[width=0.27\linewidth]{images/sdd.png}}\qquad
\subfloat[]{\includegraphics[width=0.45\linewidth]{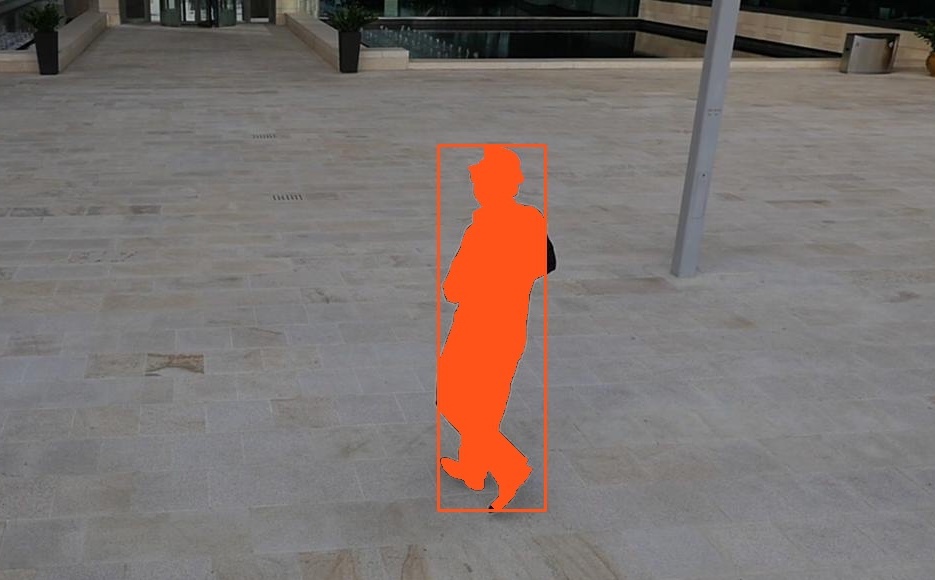}}\qquad
\subfloat[]{\includegraphics[width=0.45\linewidth]{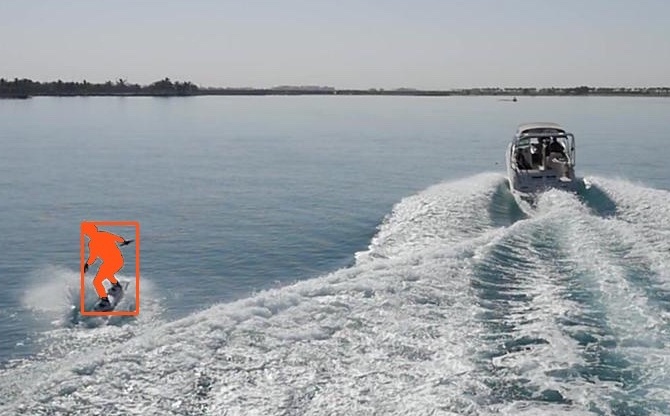}}\qquad
\subfloat[]{\includegraphics[width=0.45\linewidth]{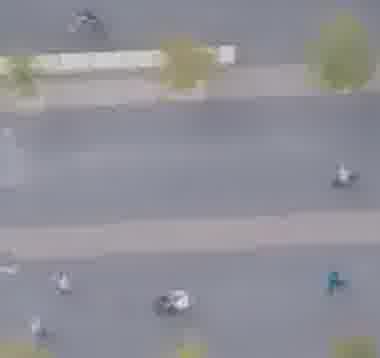}}\qquad
\subfloat[]{\includegraphics[width=0.45\linewidth]{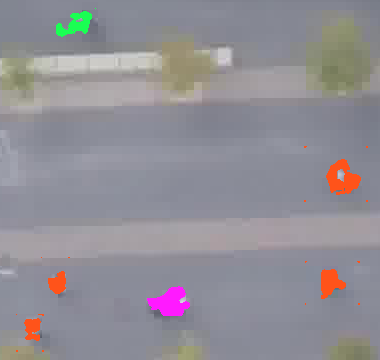}}\qquad
\caption{Instance Segmentation results from UAV123 Dataset \cite{mueller2016benchmark} and Stanford Drone Dataset\cite{sdd}.  Subfigure (a) is from `person18' video and Subfigure (b) is from the `wakeboard1' video of UAV123 Dataset. The object detection bounding box is also shown for reference. The bottom row contains the input frame from the Bookstore Scene of SDD (left) and its corresponding segmentation result (right). Red color annotates pedestrians, while violet and green colors annotate cart and biker respectively.}
\label{fig:inst_segmentation}
\end{figure}

In order to showcase the fact that \method generates instance segmentation results for datasets that contains just bounding box annotations and not any instance segmentation ground truths, we choose Stanford Drone Dataset (SDD) \cite{sdd} and UAV123 Dataset \cite{mueller2016benchmark} to carry out our experimentation. SDD contains videos shot from a drone, where each frame is annotated with objects of six categories. 
The objects in SDD are very small as the dataset contains aerial view of objects from high altitude. The UAV123 dataset contains objects of standard sizes. 
% While SDD contains high altitude drone footages, UAV123 contains low altitude footages.
A bounding box is drawn inscribing each of the object in the frames. We train the R-FCN \cite{rfcn} object detector on the datasets. Each of the detections of R-FCN is passed through the methodology described above to generate instance segmentation.
Qualitative results for the instance segmentation from UAV123 and SDD datasets are shown in Figure \ref{fig:inst_segmentation}. No quantitative results could be presented since these datasets do not have segmentation ground truth. %The red rectangle is the detection, while the segmentation can be seen as the red blob. 

%%%%%%%%% FIGURE STARTS
%%%%%%%%% FIGURE ENDS
% %%%%%%%%% FIGURE STARTS
% \begin{figure}
% \centering
% % \subfloat[]{\includegraphics[width=0.27\linewidth]{images/sdd.png}}\qquad
% \subfloat[]{\includegraphics[width=0.45\linewidth]{images/bookstore.jpg}}\qquad
% \subfloat[]{\includegraphics[width=0.45\linewidth]{images/bookstore_seg.png}}\qquad
% \caption{Instance segmentation results on a frame from Bookstore scene in Stanford Drone Dataset\cite{sdd}. Red color annotates pedestrians, while violet and green colors annotate cart and biker respectively. (best viewed in color).}
% \label{fig:sdd_detections}
% \end{figure}
% %%%%%%%%% FIGURE ENDS
\subsection{Improving Quality of Detection Datasets using \method}\label{sec:ImprovingDataset}
The availability of large amounts of high quality annotation is a prerequisite for any deep learning model to work. The success of deep learning models in image classification, object detection, segmentation and many other computer vision task has achieved impressive performance just because of the availability of quality annotations from datasets like ImageNet \cite{imagenet_cvpr09}, Pascal VOC \cite{pascal-voc-2012}, MS COCO \cite{coco}, Cityscapes \cite{cityscapes}, etc. Creating such datasets with quality annotations is a costly and time consuming activity. It limits the applicability of proven deep learning techniques to a new domain. An example would be to adapt a semantic segmentation model for understanding urban scenes, trained on CityScapes \cite{cityscapes} to some place outside Europe, or detecting objects from aerial footages using standard object detection techniques like Faster-RCNN \cite{fasterrcnn} or R-FCN \cite{rfcn}. 

One common way to alleviate the efforts to create annotated datasets, especially in videos, is to manually annotate certain key frames and then interpolate the annotation to few successive frames. This can lead to spurious annotations especially when the target under observation becomes occluded or exits the frame. Given the volume of the data that needs to be annotated, such false positives often creeps into the final dataset due to manual error. \method can provide an efficient and automated solution to filter out annotations that actually contains no object inside its annotated area, or refine the annotations.

Most of the annotations for object detection involve the coordinates of rectangular bounding boxes that enclose the object. Such rectangular annotations inherently gives scope for the background of the object under consideration to be part of the annotated area. This concern gets amplified when we consider labeling small objects, which occupies significantly small amount of pixels in an image. Practical application that demands such small objects to be annotated would be to study migration statistics of birds by detection them from videos footages by ornithologists, detecting ground objects from drone footages for surveillance activities, etc. Creating precise bounding boxes that wraps around just the object under consideration is an extremely difficult task and at most times we might have to settle down for lower quality annotations. \method can be used to enhance such annotations by making them wrap around the object under consideration much closely.

\method can be used effectively for solving both the issues enumerated above by using the following methodology. For each of the low quality annotation in such datasets, \method can generate the Objectness heatmap as explained earlier in Section \ref{sec:ObjectnessFramework}. If the pixel intensities in those heat maps are below a specific threshold, it essentially means that there is no dominant foreground object in the annotated region under consideration. Hence, those annotations can be removed from the dataset. This helps to reduce false positive annotations. 
For those heatmaps with significant intensity values, we can obtain tighter bounding boxes by considering the box inscribing the largest contour of the heatmap.

\begin{figure}[h] \centering
    \subfloat[Ground Truth annotation form Stanford Drone Dataset \cite{sdd}]
    {\includegraphics[width=0.95\textwidth]{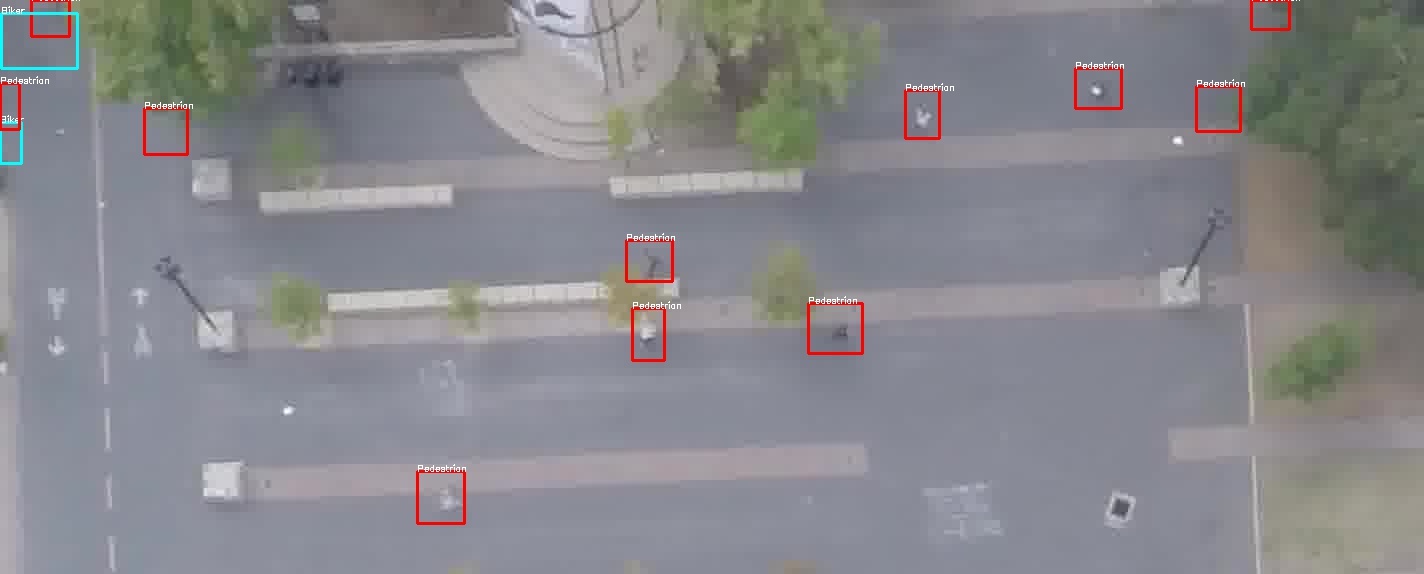}}
    \hspace{2mm}
    \subfloat[Results after using \method to remove false positive annotations and making the bounding boxes tight.]
    {\includegraphics[width=0.95\textwidth]{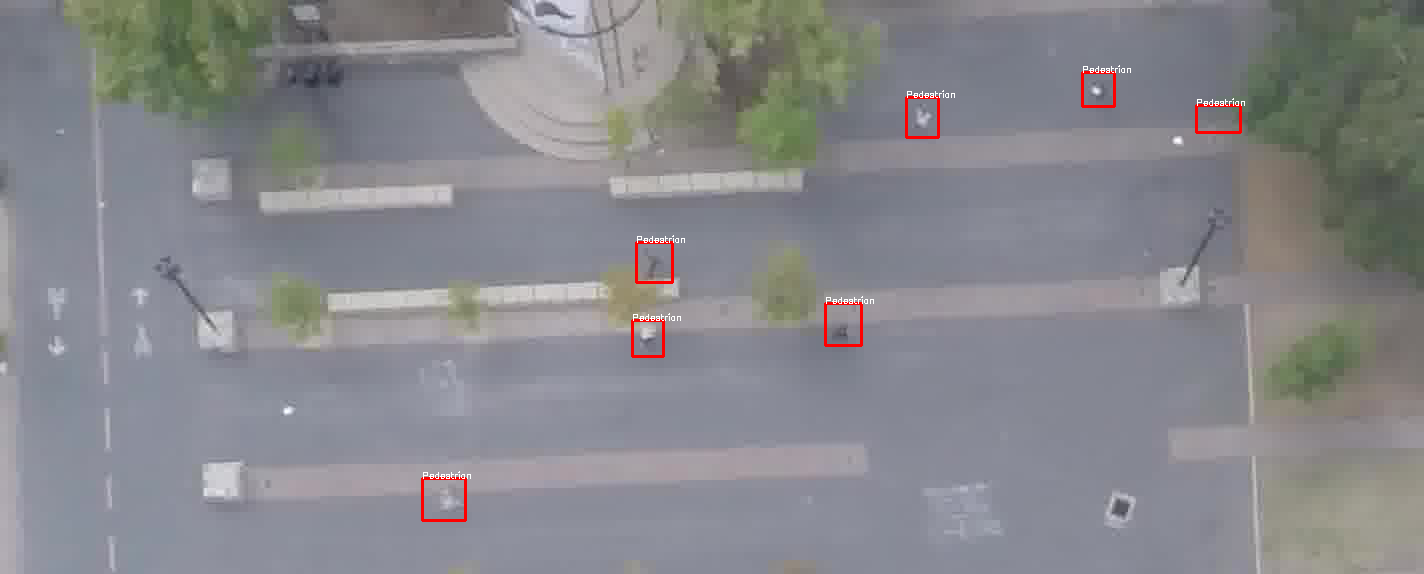}}
    \caption[]{The picture captures the effectiveness of \method to enhance bad annotations in a dataset. False positive annotations are removed and remaining annotations are made tighter by \method. The image is the 4005th frame from Video 0 of `Bookstore' scene in the Stanford Drone Dataset \cite{sdd}. Red color bounding boxes annotate Pedestrians while blue color annotate Biker.}
    \label{fig:stanforddataset}
\end{figure}

We study the usefulness of \method on this task using the Stanford Drone Dataset (SDD) \cite{sdd}, 
which contains annotated UAV footages from eight different parts of Stanford University Campus. Each of the objects are very small, relative to the size of the image. Bicyclists, pedestrian, skateboarders, carts, cars and buses are annotated with rectangular bounding boxes. This dataset was introduced in ECCV 2016, with the primary focus on accelerating research in trajectory prediction of objects. Figure \ref{fig:stanforddataset} (a) contains a sample frame from the dataset.

Human annotators have annotated some of the frames and the successive frames were automatically extrapolated. Hence these annotations are not very accurate as is evident from  Figure \ref{fig:stanforddataset} (a). The authors of the dataset included an explicit meta-data on which all annotations were manually labeled. Out of 361129 annotated objects in Video 0 of `Bookstore' scene, 6036 objects were manually annotated and 355093 objects were interpolated. The \method based methodology that is explained above can help filter out bad annotations and make the existing bounding box tighter, adding value to any task that the dataset is intended for. Figure \ref{fig:stanforddataset} (b) shows the result after applying \method based methodology to sub-figure \ref{fig:stanforddataset} (a).
% Horizontal Image STARTS
% Horizontal Image ENDS
After cleaning up the annotations from the Bookstore Scene of Stanford Drone Dataset, an R-FCN \cite{rfcn} object detector is tasked to learn a model for object detection. The RFCN is trained with frames from the `Bookstore' scene on an NVIDIA P-100 GPU with two images per mini-batch. A learning rate of 0.001 with a weight decay of 0.0005 is used. The momentum is fixed to 0.9. 
Frames from `DeathCircle' and `Hyang' scenes are used for validating the trained model. Visual characteristics like lighting and background is strikingly different in the scene used for training and testing. Hence, the generalization capability of the model is put to test. 

Table \ref{ap_of_objDetector} shows the mean average precision (mAP) of the R-FCN object detector which was trained using original annotations from the dataset and the annotations that were cleansed by removing the bad annotations. \method based pre-processing is able to improve the detection accuracy by two folds as is captured by the last two columns of Table \ref{ap_of_objDetector}.
To the best of our knowledge, no other work has been published in the research community, providing baseline for object detection on Stanford Drone Dataset. Hence this result will stand as the current baseline.

\begin{table}[h]
\centering
\begin{tabular}{|c|c|c|c|c|}
\hline
\multirow{2}{*}{Class} & \multicolumn{2}{c|}{Without Enhancement} & \multicolumn{2}{c|}{With Enhanced Dataset} \\ \cline{2-5} 
                       & AP @ 0.5            & AP @ 0.7            & AP @ 0.5           & AP @ 0.7          \\ \hline\hline
Pedestrian             & 0.302                    &    0.093                 & 0.763                   & 0.542             \\
Biker                  & 0.303                    &    0.124                 & 0.711                   & 0.521             \\
Skater                 & 0.144                    &     0.092                & 0.589                  & 0.436             \\
Car                    & 0.330                    &     0.228                & 0.554                  & 0.440             \\
Bus                    & 0.241                    &    0.282                 & 0.591                   & 0.553             \\
Cart                   & 0.577                    &    0.420                 & 0.750                   & 0.684             \\ \hline \hline
mAP                    &  0.316                   &    0.207                 & 0.659                   & 0.529             \\ \hline
\end{tabular}
\caption{
Table shows the mAP values of object detection on SDD 
using R-FCN object detector
, with and without using enhanced annotations.
}
\label{ap_of_objDetector}
\vspace{-2em}
\end{table}

\section{Discussions} \label{sec:Discussion}

\subsection{Object Proposal Generation using \method}\label{sec:ComparisonWithOPM}
\begin{figure}[th]
\begin{center}
\includegraphics[width=0.60\linewidth]{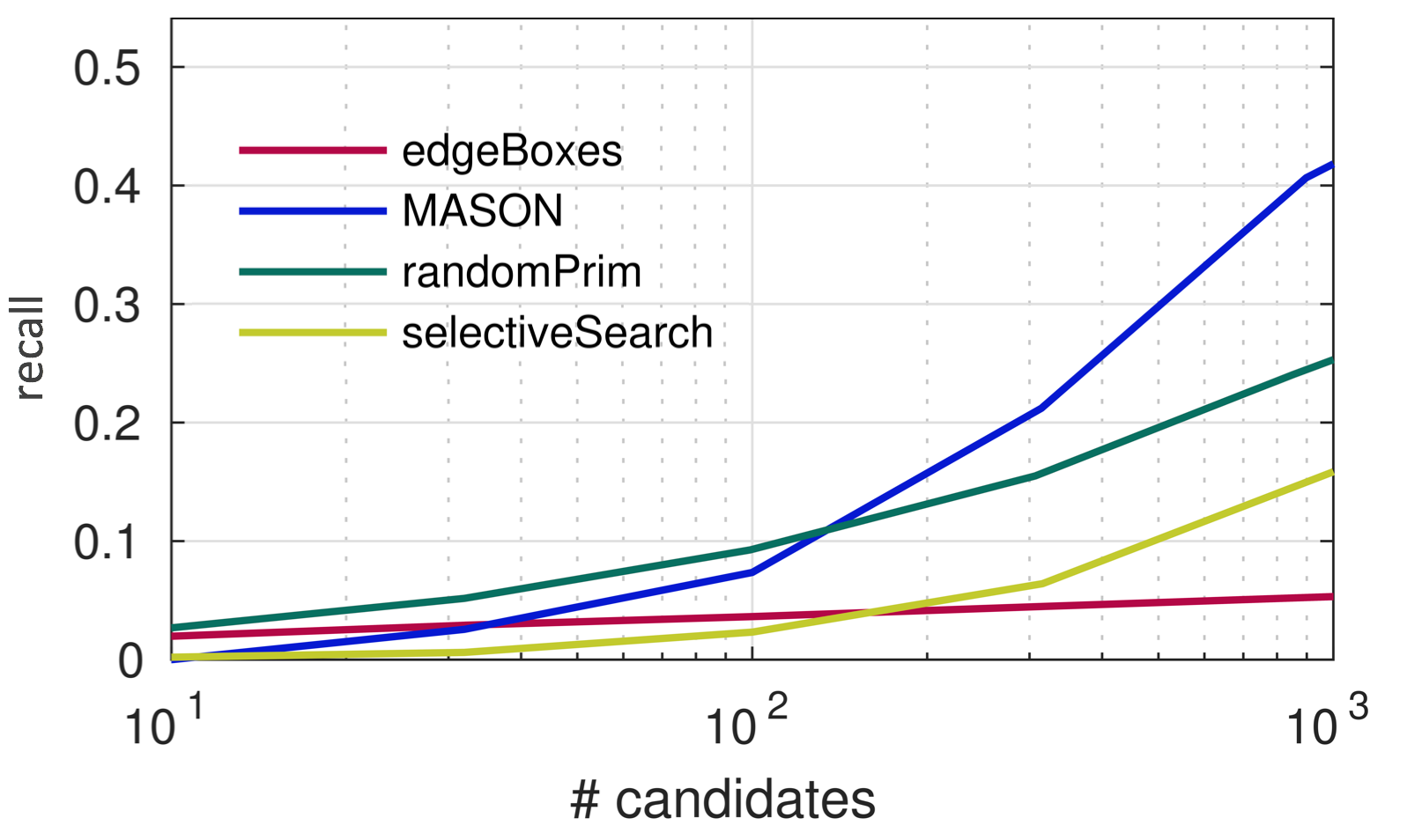}
\end{center}
\vspace{-12pt}
   \caption{Comparison of Recall of \method with the other object proposal methods.}
   \vspace{-12pt}

\label{fig_recall}
\end{figure}

As the objectness heatmap that is generated by \method is class agnostic, it can be used to generate region proposals. Multi scale boxes around blobs in heatmap is considered as proposals from \method.
In order to validate its capability, we compare against three best-in-class object proposal methods: Edge Boxes \cite{EdgeBoxes}, Selective Search \cite{SelectiveSearch} and Randomized Prim's \cite{RandomizedPrims}. All these methods are evaluated on 4952 images from the Pascal VOC 2007 \cite{pascal-voc-2012} test set. 
We used the framework proposed by Chavali \etal \cite{chavali2016object} to compare the performance of each of the object detectors. 
The average recall @ 0.9 of each of the detector is captured in Figure \ref{fig_recall}, where \method outperforms others. 
\method goes a step further to predict the pixel location of the object, which the conventional Object Proposal methods does not address. 
% This ability forms the basis for the fine-grained object localization methodology proposed in Section \ref{sec:IntegWithGrabCut}.
 
% %%%%%% FIGURE Starts
% %%%%%% FIGURE Ends

\subsection{Comparison with CAM and Grad-CAM}\label{sec:ComparisonWithCAM}
Recently, Zhou \etal \cite{zhou2016learning} proposed Class Activation Maps (CAM) using global average pooling of the activation maps to generate discriminative localization of objects in an image. Their method require changes to the architecture of standard image classification networks like VGG-16. Hence the network has to be retrained to achieve good localization results. Grad-CAM \cite{Grad-CAM} is a  modification to CAM, that can produce localization heat maps by making use of off-the-shelf image classification networks.  Hence it closely matches with \method. 

\begin{figure}
% \vspace{-1.5em}
\centering
\subfloat[Input Image]{\includegraphics[width=0.27\linewidth]{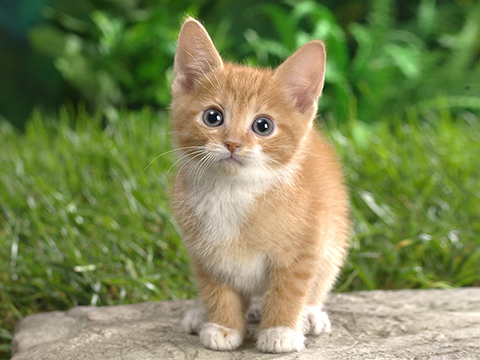}}\qquad
\subfloat[Grad-Cam HeatMap]{\includegraphics[width=0.27\linewidth]{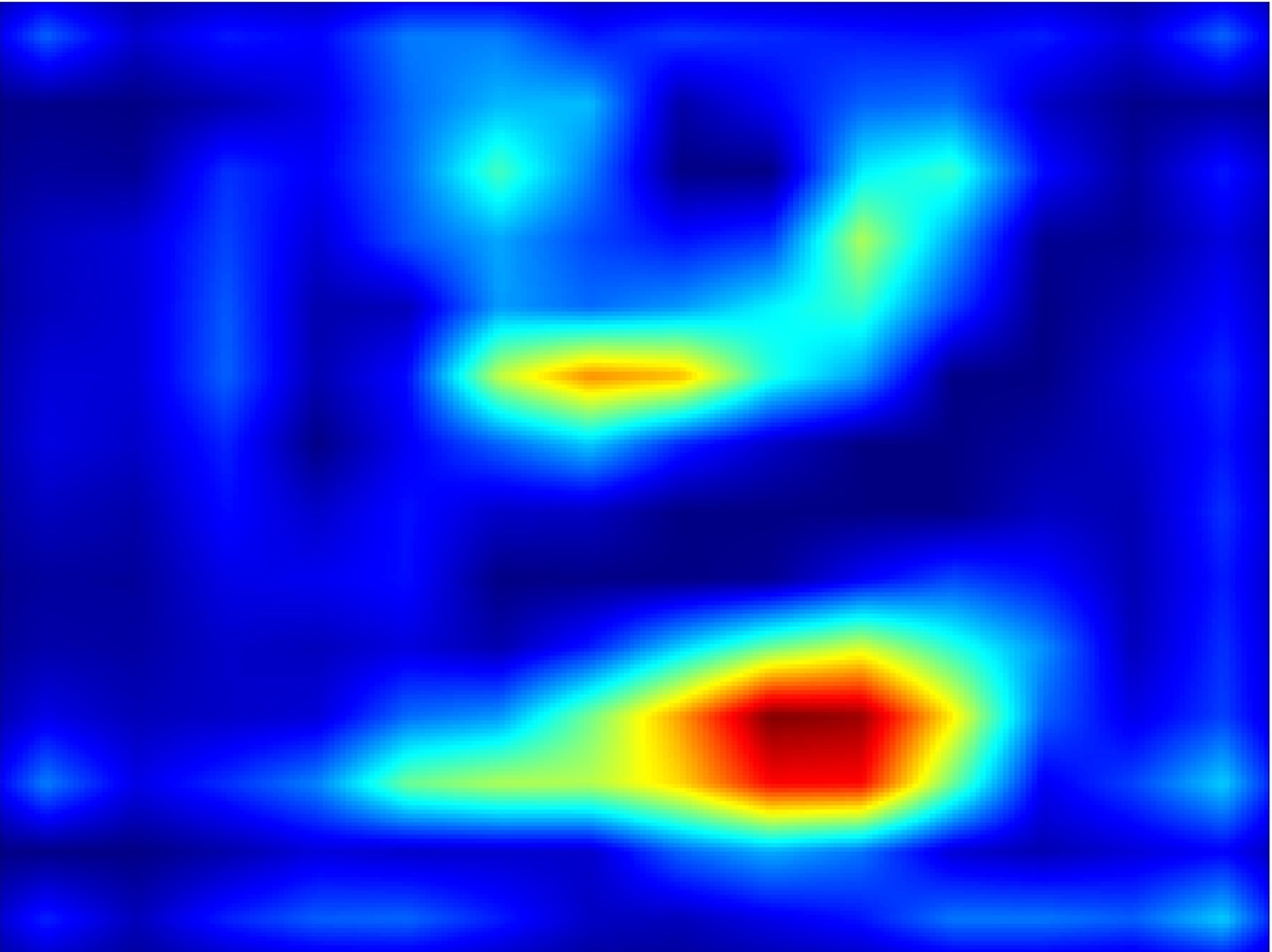}}\qquad
\subfloat[\method HeatMap]{\includegraphics[width=0.27\linewidth]{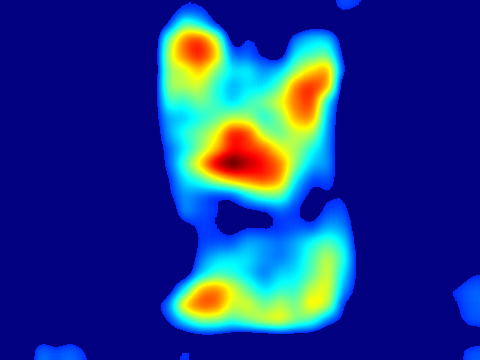}}\qquad
% \vspace{-1.5em}
\caption{A comparison of the heat maps generated by Grad-CAM \cite{Grad-CAM} and \method. Sub figure (a) is the input image and (b) and (c) are the heatmaps generated by  Grad-CAM and \method respectively. Grad-CAM HeatMap is generated for the TabbyCat class from ImageNet Dataset.
\cite{imagenet_cvpr09}.  (best viewed in color).}
\label{fig:comparison_with_gradcam}
% \vspace{-2.0em}
\end{figure}

Both CAM and Grad-CAM weighs the feature map from the last convolutional layer by a score, which is proportional to the importance of that feature map for a class of interest. Hence these methods are tightly coupled to a specific class, while \method is class agnostic. This helps our method to propose object regions, for those object classes that it has been never trained for. In another way, \method weighs all the feature maps at a specific convolutional layer equally. Despite this simplification, the method is able to give competitive results, when compared to Grad-CAM. Figure \ref{fig:comparison_with_gradcam} shows an example. 

%%%%%%%%% FIGURE STARTS
%%%%%%%%% FIGURE ENDS

% \vspace{-1.0em}
\section{Conclusion}\label{sec:Conclusion}
We have proposed a straightforward way to capitalize on the object localization information, that is available in activation map of ConvNets trained for image classification. The effectiveness of \method has been demonstrated on three significant use cases. The spatial resolution loss of the image that is incurred during the pooling operation of the ConvNet is currently compensated by simple bilinear interpolation. This can be replaced by  the `hole' algorithm proposed in \cite{deeplab}. This method is parameter-free. We open-source the code here \footnote{ https://github.com/JosephKJ/MASON}.

% \section*{Acknowledgement}\label{sec:Acknowledgement}
% \vspace{-0.5em}
% This work is funded by ANURAG, Defence Research and Development Organisation (DRDO), Government of India, through the CARS project. (Project Number: ANURAG/CSE/F121/2016-17/S16)

\bibliographystyle{splncs}
\bibliography{egbib}
\end{document}